\documentclass[11pt,a4paper]{article}

\usepackage{iftex}
\ifPDFTeX
  \usepackage[T1]{fontenc}
  \usepackage[utf8]{inputenc}
  \IfFileExists{lmodern.sty}{\usepackage{lmodern}}{}
\else
  \usepackage{fontspec}
\fi
\usepackage[margin=2.5cm]{geometry}
\usepackage[parfill]{parskip}
\usepackage{microtype}
\usepackage{xurl}
\usepackage[hidelinks]{hyperref}

\makeatletter
\renewcommand\@seccntformat[1]{\csname the#1\endcsname\csname #1@dot\endcsname\quad}
\newcommand\section@dot{.}
\newcommand\subsection@dot{}
\makeatother

\newenvironment{references}{%
  \section*{References}
  \begingroup\setlength{\parskip}{0.35em}\small
  \everypar{\hangindent=1.5em\hangafter=1}}{\par\endgroup}

\hypersetup{
  pdftitle={Translators as Invisible Teachers of AI: Copyright, Translation Memory, and the Political Economy of Linguistic Data},
  pdfauthor={Masaru Yamada}
}

\begin{document}

\begin{center}
  {\LARGE\bfseries Translators as Invisible Teachers of AI\par}
  \vspace{0.6em}
  {\large Copyright, Translation Memory, and the Political Economy of Linguistic Data\par}
  \vspace{1.8em}
  {\large Masaru Yamada\par}
  \vspace{0.4em}
  {\small College and Graduate School of Intercultural Communication, Rikkyo University\par}
  {\small\texttt{masaru.yamada@rikkyo.ac.jp}\par}
\end{center}

\vspace{1em}

\begin{abstract}
This paper examines how the labour of translators has been transformed into foundational data capital for the age of artificial intelligence (AI). Translation memories (TM) and parallel corpora preserve a one-to-one correspondence between source and target text and therefore constitute extraordinarily valuable supervised training data for machine translation. The development of statistical machine translation (SMT), neural machine translation (NMT), the Transformer architecture, and multilingual large language models (LLMs) cannot be disentangled from the accumulation of such translation data. And yet, translators' renditions have been bought as deliverables under contract, segmented as technical objects, and processed as ``information analysis'' data under copyright law---losing, in the process, their moral, creative, and economic attribution to the translators who produced them. The paper develops two concepts to capture this process. The first is \emph{appropriation without consumption}: a mode of use in which works are not read, viewed, or listened to, but only mined for statistical features---a use that is legitimated under Article 30-4 of the Japanese Copyright Act. The second is the \emph{invisible teacherisation} of translators: the process by which translators, through the construction of translation memories, post-editing, and quality assessment, have functioned as teachers of AI without recognition as such. Drawing on the data supply chain that runs from translators through language service providers (LSPs) and platforms to model developers, on a comparative reading of Japanese, European, and United States legal frameworks, on the distinction between open and proprietary AI models, and on the premium status that human-generated data has acquired in the era of model collapse, the paper asks what translators are actually afraid of, and points toward concrete directions for redistributive design.

\medskip
\noindent\textbf{Keywords:} translation memory; parallel corpora; large language models; copyright; Article 30-4 of the Japanese Copyright Act; appropriation without consumption; data labour; translation ethics; Transformer.
\end{abstract}

\section{Introduction: What Are Translators Afraid Of?}

The rapid spread of generative AI has produced a complex emotional response among professional translators. It is not merely occupational anxiety. What translators fear is not only that their work will be replaced by machines, but---more fundamentally---the suspicion that the translations, term choices, stylistic decisions, revision histories, and post-edited outputs they have produced over years have quietly become the data capital used to train AI systems, and that those systems are now eroding the very market value of the translators themselves.

This sense is no longer a matter of personal impression. It is documented by professional associations. In a survey of its members and other authors conducted in January 2024, the Society of Authors (2024) found that more than one third of the translators who responded had lost work due to generative AI, and that more than 40 percent reported a decrease in the value of their income. The International Federation of Translators (FIT, 2025), in its position paper on machine translation in the age of AI, notes that the use of training data harvested from public content raises questions about copyright, and calls for transparency in how data are used and for respect for intellectual property. The ``fear'' of translators is not merely an emotional reaction; it is grounded in reported changes in work and income.

This paper begins by decomposing this composite fear into four layers. The first is the \emph{fear of replacement}: the direct anxiety that one's job will be taken over by AI. The second is the \emph{fear of imitation}: the worry that one's style, lexical preferences, and domain expertise will be reproduced without one's participation. The third is the \emph{fear of unpaid training}: the protest that one's prior work has been used for AI development without permission, remuneration, or recognition. The fourth, and most fundamental, is the \emph{ontological fear}: the dread that translation will be redefined from ``creative judgement'' into ``replaceable data processing,'' dissolving the profession itself. The present paper focuses on this fourth layer---the ontological re-description of the profession.

Translators have not been suddenly threatened by AI. Translators have been teaching AI for a long time. What has been missing is recognition of that pedagogical role. Translation memories were introduced as productivity tools, but they were also devices that aligned source and target segments, accumulated translator judgements in machine-readable form, and converted them into reusable supervised data. Across the histories of SMT, NMT, and multilingual LLMs, parallel corpora and translation data have played a central role. The Transformer, the foundational architecture of contemporary LLMs, was likewise developed and benchmarked in the arena of machine translation (Vaswani et al., 2017).

That said, the development of LLMs cannot be reduced solely to translation data. The first GPT was framed as a combination of generative pre-training on large unlabelled corpora and task-specific fine-tuning (Radford et al., 2018). It would therefore be wrong to claim that ``LLMs were born of translation data alone.'' But across the history of machine translation, multilingual representation learning, the Transformer, parallel-corpus exploitation, and instruction tuning for translation, it is hard to deny that translation data have been a foundational, not a peripheral, resource for language AI (Lin et al., 2025).

The goal of this paper is not to mount a defence of the profession by claiming that translators deserve exceptional protection. Rather, through the experience of translators, it asks how copyright, contracts, remuneration, recognition, and the distribution of public benefits should be redesigned once human intellectual labour has become the infrastructure of AI. Section~2 re-reads the structure of translation memory as supervised training data. Section~3 traces the place of translation data in the technical history from machine translation through the Transformer to LLMs. Section~4 decomposes the data supply chain into four tiers running from translators through LSPs and platforms to model developers. Section~5 analyses the mismatch between translation and copyright. Section~6 develops the notion of appropriation without consumption under Article 30-4 of the Japanese Copyright Act, in comparison with the European and United States legal frameworks. Section~7 introduces the distinction between open and proprietary AI and discusses the premium status of human-generated data in the era of model collapse. Section~8 extends the argument to other forms of knowledge labour. Section~9 concludes by proposing concrete directions for redistributive design.

\section{Translation Memory as Supervised Data}

\subsection{The Dual Character of Translation Memory}

A translation memory (TM) is a database that aligns source-language segments with target-language segments. In day-to-day practice, it is a tool for terminological consistency, quality control, and productivity. From the standpoint of machine learning, however, a TM is a data structure that stores massive amounts of input--output correspondences: source text as input, target text as desired output---a format ideally suited to supervised learning.

Seen in this light, translation memory was never merely a productivity aid. It was an institutional and technical infrastructure that converted translators' judgements into reusable paired data. Translators thought they were rendering texts; the industry segmented those renderings, accumulated them, matched them, discounted them via fuzzy-match rates, and reused them. A translation is no longer a one-off deliverable; it has become a resource available for future translation, machine translation, quality estimation, post-editing, and, ultimately, AI training.

\subsection{A Reuse the Berne Convention Did Not Anticipate}

Moorkens and Lewis (2020) point out that the copyright framework envisioned by the Berne Convention did not anticipate large-scale translation reuse as TM content, much less as MT training data. Modern copyright was conceived around the reproduction, publication, and sale of works; the AI era, by contrast, foregrounds a different mode of use: extracting statistical features from works rather than ``letting people read them.'' This qualitative shift in the nature of use is a hard problem for existing copyright theory.

The discussion of TM ownership exemplifies the same structural shift. Topping (2000) had already asked whether translation memories should belong to the end client, the language service provider, or the translator (as cited in Moorkens \& Lewis, 2020). In practice, clients who commission and pay for a translation tend to assume that they ``own'' it, and TM files are usually delivered to them together with the translated files, while the copyright potentially attributable to the translator is hardly ever asserted (Moorkens \& Lewis, 2020). Yet from the translator's side, this sense of ``ownership'' tacitly absorbs into the original translation fee a long list of future entitlements: rights of reuse, rights of use for machine learning, and rights to extract stylistic value.

\subsection{Support Tool, or Infrastructure of Extraction?}

The evaluation of TM must remain ambivalent. TMs have been genuinely useful to translators: tools for terminological consistency, throughput, and quality assurance, supporting a cognitively demanding activity. But they have simultaneously been the institutional device by which clients and language service providers (LSPs) accumulate and reuse translators' past labour, and through fuzzy-match-rate discounting, continuously suppress translator pay. Moorkens (2020) argues that the translation industry exhibits a form of ``digital Taylorism,'' in which large translation companies break work into smaller tasks and rigidly define and monitor translation processes, particularly in platform-mediated workflows.

Translation memory, in short, was at once a productivity tool for translators and a device for converting their judgements into reusable, machine-readable data. This dual character is the institutional origin of the anxiety translators experience in the present AI era.

\section{Translation as a Laboratory for AI}

\subsection{From SMT to NMT}

Machine translation has long been a central problem in natural language processing. SMT relied on probabilistic models trained over parallel corpora; NMT, similarly, was trained on source--target pairs. The attention-based NMT proposed by Bahdanau, Cho, and Bengio (2015) introduced a mechanism by which a model learns where in the source to focus when generating translation, and it achieved performance comparable to existing phrase-based methods in English--French translation. The Google Neural Machine Translation system (Wu et al., 2016) then positioned NMT as end-to-end learning of translation, demonstrating large-scale quality gains in a deployed production system.

\subsection{The Transformer Was Born of Translation}

What deserves emphasis here is that the Transformer---the foundational architecture of contemporary LLMs---emerged from the machine translation task. The ``T'' in GPT stands not for Translator but for Transformer; yet the Transformer itself was designed, evaluated, and disseminated within the problem of translation. \emph{Attention Is All You Need} (Vaswani et al., 2017) proposed an architecture composed solely of attention mechanisms, without recurrence or convolution, and demonstrated state-of-the-art performance on the WMT 2014 English--German and English--French translation tasks.

One can therefore put it as follows. The ``T'' in GPT does not stand for Translator. But the Transformer leapt from the problem of translation to the status of a general-purpose architecture for the world, and so it is fair to say that the labour of translators has sedimented at the very bottom of the history of generative AI. Translation was not, for AI, simply an application domain. It was the central laboratory in which language understanding, sequence-to-sequence transformation, attention, contextual representation, and multilinguality were tested.

\subsection{The Place of Translation Data in the LLM Era}

It is important not to overstate this claim. Since the first GPT (Radford et al., 2018), the core of autoregressive LLMs has been generative pre-training on large unlabelled, monolingual text, and parallel corpora are not its principal material. Yet for the strengthening of multilingual LLMs and the improvement of translation ability, parallel corpora remain an active research focus. Lin et al. (2025) systematically examine how parallel corpora can enhance multilingual LLMs, investigating the quality and quantity of parallel data, training objectives, and model size; among other findings, they report that filtering out noisy translations matters most and that relatively small amounts of parallel data can be effective.

The argument of this paper can thus be sharpened. The development of LLMs in general cannot be reduced solely to translation data. But across the history of machine translation, multilingual representation learning, the Transformer architecture, parallel-corpus exploitation, and instruction tuning for translation, translation data have been not a peripheral material but one of the foundational resources of language AI. Translators have been the suppliers of that foundational resource.

\section{The Data Supply Chain}

Translation data do not travel directly from translators to AI developers. Multiple intermediaries lie between them. A binary opposition of ``translators versus AI companies'' obscures both the pathways of extraction and the locus of legal responsibility. This section disaggregates the data supply chain into four tiers.

\subsection{From Translators to Language Service Providers (LSPs)}

The first stage is the moment at which a translator delivers a rendered text to a language service provider (LSP). In most LSP contracts, ownership and reuse rights in the translation memory are assigned to the client or to the LSP itself. Freelance translators, lacking explicit bargaining power, transfer the rights to secondary use of their translations as a matter of course. The LSP then reuses the TM in subsequent projects and discounts the translator's rate through fuzzy-match-rate calculations. The extraction at this stage is a continuation of classical labour management, and has already been characterised as digital Taylorism (Moorkens, 2020).

\subsection{From LSPs to Platforms}

The second stage is the moment at which the LSP draws on its accumulated TM to develop in-house machine translation engines, or sells the data to third parties. Some large LSPs train proprietary NMT engines on massive volumes of internal TM, in a process that directly converts translator labour into engine assets. The problem here is that what translators delivered as ``translation work'' is repurposed beyond their recognition into ``engine training''---an entirely different economic activity.

\subsection{From the Open Web to Model Developers}

The third stage is the moment at which parallel data exist on the open web and are scraped by large technology firms. Parallel corpora such as OPUS, ParaCrawl, and CCMatrix---constructed from public-sector documents, subtitles, government publications, and volunteer translation---are extensively used in multilingual pre-training of LLMs. Along this route, there is no contractual relationship, and indeed no mutual recognition, between the individual translator and the ultimate user (the model developer).

\subsection{The Continuous Supply Provided by Post-editing}

The fourth stage is the ongoing labour of post-editing. The act of correcting machine translation output appears on the surface to be ``assistance to the machine,'' but structurally it produces high-quality training data that pair the machine's errors with the human's corrections. Post-editing is the moment at which translators not only use AI but also act as teachers improving AI. Yet this pedagogical role is almost entirely unreflected in either the remuneration system or the rights system.

\subsection{Why This Disaggregation Matters}

Making this four-tier structure explicit has two payoffs. First, it disentangles the locus of legal responsibility. Contractual rights transfer at the LSP stage is a problem of contract law; web scraping is a problem of copyright limitations; post-editing is a problem of labour management and unwaged labour. Second, it clarifies the unit at which counter-strategies should be designed. Trade-union-style organising among translators is effective at the LSP stage; data governance design is effective at the platform stage; copyright reform is effective at the open-web stage. As long as the problem is treated as a monolith, counter-strategies remain blurred.

\section{From ``Expression'' to ``Data'': The Border of Copyright}

\subsection{Translation as a Derivative Work}

Copyright law protects the creative expression of thought and feeling. A translation is a derivative work that depends on a source work, and copyright can arise in the translator's creative expression of the target text. Translation therefore exhibits a three-layered structure: the rights of the original author, the rights of the translator, and the user's obligation to obtain permission from both. In commercial translation, however, the translator's economic rights in the target text are typically transferred to the client by contract. As a result, even when those translations are later used to train AI, the translator has often already lost the capacity to object as a rights holder.

\subsection{The Idea/Expression Dichotomy and ``Style''}

A foundational principle of copyright law is the idea/expression dichotomy: copyright protects concrete expressions, but not the ideas, conventions, styles, or methods that lie behind them. Applied to translation, the ``literal'' or ``free'' translation style as such, or the ``manner'' of a particular translator, is in principle not protected. The verbatim reproduction of a translator's actual rendering, by contrast, clearly falls within the scope of copyright.

This is where the structural difficulty of protecting translators in the AI era surfaces. The market value of a translator does not reside solely in the copyrightability of individual sentences. It resides in an accumulated set of judgements: feel for language, style, domain knowledge, lexical consistency, depth of reading, treatment of ambiguity, and cultural mediation---precisely the kinds of judgement that copyright is not well suited to protect. Yet AI learns exactly that hard-to-protect region. The core of a translator's value is poorly protected by law but readily extracted by technology.

The disjunction can be stated in one sentence. Copyright protects translators' concrete renderings, but not the stylistic patterns, decision tendencies, lexical preferences, and reading habits that constitute the core of their professional value. The problem of the AI era is that this poorly protected region has become the most valuable target of learning for the model.

\section{Article 30-4 of the Japanese Copyright Act and ``Appropriation Without Consumption'': A Comparative Legal Reading}

\subsection{AI Training as ``Non-enjoyment Use'' under Japanese Law}

Article 30-4 of the Japanese Copyright Act (\emph{Chosakukenh\=o dai-sanj\=u-j\=o no yon}) permits the use of copyrighted works for purposes such as information analysis, and is widely regarded internationally as one of the most permissive limitations on copyright with respect to AI training. The provision is structured around three core elements: (i) the use must not be ``for the purpose of enjoying the thoughts or sentiments'' expressed in the work; (ii) it must be ``within the scope necessary'' for that purpose; and (iii) it must not ``unreasonably prejudice the interests of the copyright holder.''

This provision has enormous significance for AI training. When a work is used not to be read, viewed, or listened to, but to be analysed statistically, the use generally falls within the limitation. Feeding translation memories or parallel corpora into the pre-training or continued pre-training of an LLM can, as a rule, be located within this ``non-enjoyment use.'' Some commentators read the provision more broadly still, arguing that machine learning involves no purpose of enjoyment even when it aims to output protected expression, and that similar outputs should be addressed at the stage of use rather than training (Matsushita, 2025).

\subsection{The Structural Inversion of ``Appropriation Without Consumption''}

From the translator's standpoint, however, there is a paradox. What translators fear is not only that their translations will be read without permission. It is that their translations will be analysed in a form in which they are not enjoyed, and that translation models or style-imitating models will be built from that analysis. Classical copyright infringement has been organised around unlicensed enjoyment of a work. In AI training, by contrast, the use is legitimated because it is not enjoyment.

The present paper terms this structural inversion \textbf{appropriation without consumption}. The value of a translator's rendering is extracted not by being read by readers, but by being analysed without any reader. This is the fundamental shift in copyright theory required by the AI era. Where prior infringement was constituted along the consumption pathway---being read, being sold---contemporary appropriation proceeds by routing around that pathway.

\subsection{The European Union: Opt-out Rights and Transparency Obligations}

The legal architecture of the European Union takes the opposite design from the Japanese one. Article~4 of the Directive on Copyright in the Digital Single Market (DSM Directive, 2019) permits commercial text and data mining (TDM) only insofar as rights holders have not expressly reserved the use of their works in an appropriate manner---by machine-readable means in the case of content made publicly available online. The default in the EU, in other words, is that rights holders retain an opt-out right. The EU Artificial Intelligence Act (AI Act, 2024) further requires providers of general-purpose AI models to put in place a policy to comply with such reservations and to publish a ``sufficiently detailed summary'' of the content used for training. This is an institutional design intended to give rights holders---including translators who hold rights in their work---the basis on which to infer whether their works have been used in training.

\subsection{The United States: Fair Use and Judicial Determination}

The United States has no statutory copyright limitation directly governing TDM or AI training, and relies instead on the fair-use doctrine. \emph{Authors Guild v.\ Google} (2015) held the large-scale scanning of Google Books to be transformative use. Courts have since begun to address AI training directly, with divergent results. In June 2025, two judges of the Northern District of California found the training of large language models on books to be fair use (\emph{Bartz v.\ Anthropic}, 2025; \emph{Kadrey v.\ Meta}, 2025), although the \emph{Bartz} court held that retaining pirated copies in a central library was not fair use, and the \emph{Kadrey} ruling was expressly limited to the record before the court. By contrast, the District of Delaware rejected a fair-use defence for training a non-generative legal research tool on Westlaw headnotes, and the Third Circuit affirmed in September 2026 (\emph{Thomson Reuters v.\ Ross}, 2025). Other cases, including \emph{Andersen v.\ Stability AI} and \emph{The New York Times v.\ OpenAI}, remain pending at the time of writing. Because the United States framework relies on case-by-case judicial determination, its outcomes remain highly uncertain for rights holders and create significant litigation risk for developers.

\subsection{Why Japanese Law May Be Especially Extractive}

Placed side by side, Article 30-4 of the Japanese Copyright Act exhibits a distinctive profile. Whereas the EU institutionalises both an opt-out right and a transparency obligation, and the United States retains an ex-post judicial check through fair-use litigation, Japan permits non-enjoyment uses on an ex-ante, comprehensive basis, and provides neither a statutory opt-out right nor a transparency obligation. As a consequence, under Japanese law a translator has, in institutional terms, virtually no means of confirming whether her work has been used in training, and virtually no means of refusing such use. In this sense, the Japanese regime can be described as structurally ``especially extractive'' when set against the global comparison.

\subsection{The Proviso and Difficulties of Proof}

Article 30-4 contains a proviso that excludes uses which ``unreasonably prejudice the interests of the copyright holder.'' The guidance of the Legal Subcommittee of the Copyright Subdivision of the Cultural Council (2024) also identifies limits to the provision. Style as such is treated as an idea and is not protected; but additional training on a small set of works by a specific creator, intended to output the creative expression shared by those works, is treated as having a purpose of enjoyment, so that Article 30-4 does not apply. The examples the guidance gives for the proviso itself concern markets for databases compiled for information analysis. There is, in principle, room to apply similar reasoning to targeted training on the renderings of a particular translator.

In practice, however, the obstacles are formidable. First, only the rights holder has standing to sue. A translator who has already assigned rights under a standard commercial-translation contract is unlikely to be the proper claimant. Second, proving that a specific translation was included in training data is technically and procedurally extremely difficult under existing discovery regimes. Third, no settled benchmark exists for when the imitation of a translator's renderings moves from shared style to shared creative expression, or when it amounts to ``unreasonable prejudice.'' The net effect is that Article 30-4 risks legally ratifying, rather than constraining, the loss of translators' rights.

\section{Open Models, Proprietary AI, and the Premium Value of Human Data in the Era of Model Collapse}

\subsection{The Ambivalence of Public Benefit and Private Revenue}

The argument of this paper is not a rejection of AI development as such. Machine translation has expanded linguistic access, narrowed information gaps, and delivered public benefit in disaster response, healthcare, education, international exchange, and migration and refugee support. That translation data have contributed to the development of AI is a result of meaningful value for humanity at large.

But the existence of public benefit does not justify the unwaged invisibilisation of the labour on which that benefit rests. The question is not whether AI has progressed. The question is whose labour was used to make it progress, who was paid, who lost rights, and who has been pushed out of the market---a question of distribution.

\subsection{Are Open Models and Proprietary AI Really the Same?}

A distinction by the character of the AI model is critical here. Translation models and multilingual models built for public-interest purposes by academic institutions and non-profits---No Language Left Behind, the Helsinki-NLP OPUS-MT models---differ qualitatively in their consequences from the proprietary commercial LLMs produced by OpenAI, Google, Anthropic, and others, even when a translator's renditions enter both via the same data pathway.

Open models release their training outputs publicly; anyone can download and use them, including speakers of low-resource languages, humanitarian agencies, and educational institutions. Here, translator labour can plausibly be described as having been contributed to a ``data commons.'' Proprietary AI, by contrast, encloses training outputs as a source of competitive advantage and monetises them through API fees and subscriptions. Here, translator labour is transformed into private capital.

The core of the fear of unpaid training is often located precisely in the latter case. When translators want to draw a line between ``for humanity'' and ``for the exclusive revenue of corporations,'' this is not an emotional reaction but a legitimate logical demand for the distinction between a data commons and private capital. When this paper speaks of moving ``beyond public good without redistribution,'' it is precisely the absence of this distinction that it has in view.

\subsection{Model Collapse and the Premium Value of Human Data}

A major development in recent AI research is the phenomenon known as ``model collapse.'' Shumailov et al.\ (2024, \emph{Nature}) showed that recursively training successive generations of models on AI-generated text causes the model distribution to degrade, with rare cases and tail expressions vanishing first. As the open web becomes saturated with AI-generated content, the scarcity and value of human-generated, uncontaminated, high-quality data are, paradoxically, on the rise for AI developers.

This shifts the framing of translator labour in an important way. So far this paper has argued that translator labour has been ``appropriated as something past.'' More precisely, however, the high-quality parallel data still being produced by professional human translators---edited, proofread, and quality-controlled---has become, in the era of model collapse, a premium asset that AI companies urgently want. The source--target pair format is the ideal form of supervised data. Domain-specific terminology, style, and cultural mediation, moreover, are scarce resources that web scraping cannot readily yield.

Translators are simultaneously being appropriated for their past labour and continuing to produce premium assets in real time. It is this temporal duality that strengthens the demand for redistribution. As suppliers of a scarce resource, translators retain, in principle, some bargaining leverage. The problem is that no institutional channel currently exists for exercising it.

\subsection{The Quadruple Position of Translators and the Core of Their Fear}

Translators are simultaneously users of AI, teachers of AI, objects of replacement by AI, and---in the era of model collapse---suppliers of premium data on which AI's continued integrity depends. This fourfold position is the core of translator anxiety. How to design the tension between data commons and data extraction, between public good and private capital, and between past and ongoing labour, is one of the central institutional questions of the AI era.

\section{The Case of Translators Is Not Unique}

The translator's experience is not specific to the translation industry. Translators are, rather, a profession that has experienced earlier than most the cycle in which intellectual labour is turned into data, absorbed by AI, and returned to the labour market. The same structure extends across creative and knowledge work.

The styles of illustrators, the motifs of composers, the contract-review judgement of lawyers, the diagnoses of physicians, the corrections of teachers, the code of programmers, the editorial revisions of editors: each is treated as a discrete deliverable under copyright or contract, but to AI they are sets of judgement patterns. Human expertise is converted, inside the dataset, into statistical features.

The translation case is unusually transparent. There is a clear input--output correspondence between source and target; quality is measurable; the structure scales easily across language pairs. That is precisely why translation data has been ideal supervised data for AI. The experience of translators is a leading case of what every specialist profession will eventually face. The anxiety translators feel today is a prefigurative form of an anxiety that will eventually be felt across the knowledge professions.

\section{Conclusion: Toward Redistributive Design}

Translators are not the enemies of AI. Translators have been, rather, the invisible teachers of AI. The renditions, translation memories, parallel corpora, and post-edited outputs translators have produced over years have been deeply involved in the development of language AI. But that contribution has been bought as a deliverable under contract, segmented technically, processed legally as information-analysis data, and economically redistributed almost not at all.

The question this paper has pursued is not whether AI development should be stopped. The question is whether, given that AI development depends so deeply on human intellectual labour, it is acceptable for that labour to remain invisibilised and severed from rights and remuneration.

What translators fear is not merely losing work. It is a strange form of immortality: their labour loses its name, its rights, and its pay, but persists inside the model. Justice in the AI era begins with how we relate not only to the copyright in AI outputs, but also to the past labour that made those outputs possible.

The argument can be put most sharply as follows. Translators are not only being replaced by AI. They are being replaced by an AI that has been trained on their own past labour. That circularity is the core of the translator's fear in the era of generative AI. And the problem is not that translators' renderings have been read; it is that they have been analysed without being read. Appropriation without consumption, far from reducing the risk of infringement, has become a new pathway for extracting the professional value of translators.

\subsection*{Directions for Redistributive Design}

Rather than rest at an abstract call for ``redistribution,'' this paper closes with four concrete directions.

\textbf{First, explicit agreement at the level of contracts and collective bargaining.} Use of translations for AI training should be addressed in standard translation-services contracts, with purpose-specific consent clauses, explicit scope of reuse, and remuneration provisions. The development and dissemination of model contracts by professional bodies---FIT, national translators' associations, the Society of Authors, and others---is the practical route here.

\textbf{Second, collective rights management through data trusts.} Negotiating individually with AI companies is not realistic for freelance translators. Drawing on the experience of collective management organisations in music (JASRAC, ASCAP, etc.), intermediary bodies that aggregate and distribute revenues from translation memories and post-editing data offer a plausible institutional alternative.

\textbf{Third, technical opt-out metadata.} Machine-readable metadata indicating whether AI training is permitted on a given translation output---standards such as C2PA, the TDM Reservation Protocol, and ai.txt---should become a routine attachment to deliverables. The opt-out right enshrined in EU law cannot function unless this technical infrastructure is in place.

\textbf{Fourth, reform of Article 30-4 of the Japanese Copyright Act.} Against the structure under which appropriation without consumption is legally ratified, three legislative and administrative tasks should be placed on the agenda: (i) the introduction of a transparency obligation, (ii) the statutory recognition of an opt-out right, and (iii) the clarification of interpretive guidelines for the proviso.

Translation studies now finds itself at a juncture at which its very object---``translation labour''---has become a touchstone for the institutional design of the AI era. Translators are a leading case for every knowledge worker in the era of generative AI; the frameworks devised here will, in due course, extend to illustrators, musicians, lawyers, physicians, teachers, and programmers, and to knowledge work in general.


\begin{references}
\emph{Authors Guild v.\ Google, Inc.}, 804 F.3d 202 (2d Cir. 2015).

Bahdanau, D., Cho, K., \& Bengio, Y. (2015). Neural machine translation by jointly learning to align and translate. In \emph{Proceedings of the 3rd International Conference on Learning Representations (ICLR 2015)}. \url{https://arxiv.org/abs/1409.0473}

\emph{Bartz v.\ Anthropic PBC}, No.\ 3:24-cv-05417-WHA (N.D. Cal. June 23, 2025) (order on fair use).

Cultural Council, Copyright Subdivision, Legal Subcommittee. (2024, March 15). \emph{AI to chosakuken ni kansuru kangaekata ni tsuite} [General understanding on AI and copyright in Japan]. Agency for Cultural Affairs. \url{https://www.bunka.go.jp/seisaku/bunkashingikai/chosakuken/pdf/94037901_01.pdf}

European Parliament and Council. (2019). Directive (EU) 2019/790 of 17 April 2019 on copyright and related rights in the Digital Single Market (DSM Directive), Articles 3--4. \emph{Official Journal of the European Union}, L 130, 92--125.

European Parliament and Council. (2024). Regulation (EU) 2024/1689 of 13 June 2024 laying down harmonised rules on artificial intelligence (AI Act), Article 53. \emph{Official Journal of the European Union}, L, 2024/1689.

International Federation of Translators (FIT). (2025, June). \emph{FIT position paper on machine translation in the age of AI}. \url{https://en.fit-ift.org/wp-content/uploads/2025/08/PDP_202506_MT_EN_FINAL.pdf}

\emph{Kadrey v.\ Meta Platforms, Inc.}, No.\ 3:23-cv-03417-VC (N.D. Cal. June 25, 2025) (order on summary judgment).

Lin, P., Martins, A. F. T., \& Sch\"utze, H. (2025). A recipe of parallel corpora exploitation for multilingual large language models. In \emph{Findings of the Association for Computational Linguistics: NAACL 2025} (pp.~4038--4050). Association for Computational Linguistics. \url{https://doi.org/10.18653/v1/2025.findings-naacl.225}

Matsushita, G. (2025). Kikai gakush\=u e no chosakukenh\=o 30-j\=o no 4 no tekiy\=o ni tsuite [On the application of Article 30-4 of the Japanese Copyright Act to machine learning]. \emph{Chizai Journal}, \emph{18}, 59--74. Nihon University.

Moorkens, J. (2020). ``A tiny cog in a large machine'': Digital Taylorism in the translation industry. \emph{Translation Spaces}, \emph{9}(1), 12--34. \url{https://doi.org/10.1075/ts.00019.moo}

Moorkens, J., \& Lewis, D. (2020). Copyright and the reuse of translation as data. In M. O'Hagan (Ed.), \emph{The Routledge handbook of translation and technology} (pp.~469--481). Routledge. \url{https://doi.org/10.4324/9781315311258-28}

Radford, A., Narasimhan, K., Salimans, T., \& Sutskever, I. (2018). \emph{Improving language understanding by generative pre-training} [Technical report]. OpenAI.

Shumailov, I., Shumaylov, Z., Zhao, Y., Papernot, N., Anderson, R., \& Gal, Y. (2024). AI models collapse when trained on recursively generated data. \emph{Nature}, \emph{631}(8022), 755--759. \url{https://doi.org/10.1038/s41586-024-07566-y}

Society of Authors. (2024, April 11). \emph{SoA survey reveals a third of translators and quarter of illustrators losing work to AI}. \url{https://societyofauthors.org/2024/04/11/soa-survey-reveals-a-third-of-translators-and-quarter-of-illustrators-losing-work-to-ai/}

\emph{Thomson Reuters Enterprise Centre GmbH v.\ Ross Intelligence Inc.}, No.\ 1:20-cv-00613-SB (D. Del. Feb. 11, 2025), \emph{aff'd} (3d Cir. 2026).

Topping, S. (2000). Shortening the translation cycle at Eastman Kodak. In R. C. Sprung (Ed.), \emph{Translating into success: Cutting-edge strategies for going multilingual in a global age} (ATA Scholarly Monograph Series, Vol.~11, pp.~111--125). John Benjamins.

Vaswani, A., Shazeer, N., Parmar, N., Uszkoreit, J., Jones, L., Gomez, A. N., Kaiser, {\L}., \& Polosukhin, I. (2017). Attention is all you need. In \emph{Advances in Neural Information Processing Systems 30} (pp.~5998--6008).

Wu, Y., Schuster, M., Chen, Z., Le, Q. V., Norouzi, M., et al. (2016). \emph{Google's neural machine translation system: Bridging the gap between human and machine translation} (arXiv:1609.08144). arXiv. \url{https://arxiv.org/abs/1609.08144}
\end{references}
\end{document}